\documentclass[conference]{IEEEtran}
\IEEEoverridecommandlockouts

\usepackage{cite}
\usepackage{amsmath,amssymb,amsfonts}
\usepackage{algorithm}
\usepackage{algpseudocode}
\usepackage{graphicx}
\usepackage{textcomp}
\usepackage{xcolor}
\def\BibTeX{{\rm B\kern-.05em{\sc i\kern-.025em b}\kern-.08em
    T\kern-.1667em\lower.7ex\hbox{E}\kern-.125emX}}

\def\bvec#1{\mbox{\boldmath $#1$}}

\begin{document}

\title{Ownership Verification of DNN Models Using White-Box Adversarial Attacks with Specified Probability Manipulation
\thanks{This study was supported by the JSPS KAKENHI(25K15225), JST SICORP (JPMJSC20C3), JST CREST (JPMJCR20D3), Japan.
}
}

\author{
\IEEEauthorblockN{Teruki Sano}
\IEEEauthorblockA{\textit{Graduate School of Information Sciences} \\
\textit{Tohoku University}\\
Sendai, Japan \\
sano.teruki.r2@dc.tohoku.ac.jp}
\and
\IEEEauthorblockN{Minoru Kuribayashi$^\dagger$ \quad Masao Sakai \quad Shuji Isobe \quad Eisuke Koizumi}
\IEEEauthorblockA{\textit{Center for Data-driven Science and Artificial Intelligence} \\
\textit{Tohoku University}\\
Sendai, Japan \\
$\dagger$kminoru@tohoku.ac.jp}
}

\maketitle

\begin{abstract}
In this paper, we propose a novel framework for ownership verification of deep neural network (DNN) models for image classification tasks. It allows verification of model identity by both the rightful owner and third party without presenting the original model. We assume a \textit{gray-box} scenario where an unauthorized user owns a model that is illegally copied from the original model, provides services in a cloud environment, and the user throws images and receives the classification results as a probability distribution of output classes. The framework applies a \textit{white-box} adversarial attack to align the output probability of a specific class to a designated value. Due to the knowledge of original model, it enables the owner to generate such adversarial examples. We propose a simple but effective adversarial attack method based on the iterative Fast Gradient Sign Method (FGSM) by introducing control parameters. Experimental results confirm the effectiveness of the identification of DNN models using adversarial attack.
\end{abstract}

\section{Introduction}
\label{sec:intro}
Artificial intelligence possesses high information processing capabilities and versatility. Among its various applications, deep neural networks (DNNs) are widely utilized, and machine learning tools are offered as cloud computing services, such as Machine Learning as a Service (MLaaS). As trained deep neural network (DNN) models take significant development costs and they have high social value, it is crucial to protect the ownership of DNN models. 

Trained models are at risk of unauthorized misappropriation of their parameters and algorithms by individuals seeking to circumvent the effort and cost required for model development. To take countermeasures against such misappropriation, DNN watermarking and fingerprinting have been investigated for the protection of intellectual property of DNN models \cite{watermark-survey,Sun2023-survey}. DNN watermarking is a technique for embedding specific information in a DNN model, such as internal structure and weight parameters, during the training phase. On the other hand, DNN fingerprinting extracts some unique model properties like decision boundaries as the fingerprint. It is less realistic to assume white-box access to the model's internal structure and weight parameters for verification. Instead, it is assumed that the watermark/fingerprint will be obtained by throwing queries to a suspected model and receiving the responses. In such a case, the watermark/fingerprint reflects the behavior of the model on the given trigger images as the backdoor. The behavior is regarded as an identifier for the ownership verification. In both cases, the identifier is extracted from the suspected model through queries and then it is compared against the constructed identifier of the original model. There are two major drawbacks in these ownership verification approaches. One is that the number of trigger images is limited as those are pre-determined during the training of DNN models. 
The other risk is that unauthorized users might recognize that they are being tested in terms of their legitimacy. The behavior of the model on the trigger images is generally peculiar compared with other inputs. Due to the characteristics of queries and responses, unauthorized users may recognize the action of ownership verification approach.

In this study, we propose a novel framework for verifying the identity of DNN models by utilizing \textit{white-box} adversarial attacks \cite{I-FGSM}, allowing both rightful owners and third parties to demonstrate model ownership. With full access to the original model, the owner can generate adversarial samples that serve as proof of ownership, without disclosing the model. On any given requests from the third party on arbitrary images and specific probability values, the owner presents those accurate adversarial samples which are extremely difficult to generate without the model. Here, we assume a \textit{gray-box} scenario where a potentially copied DNN model operates in a cloud environment under the control of an unauthorized user and outputs per-class probability distributions. 

We propose a novel white-box adversarial attack that accurately manipulates the probability of a specific target class while maintaining the original class's probability. Our method allows the owner to generate adversarial samples that adjust class probabilities to designated values without revealing the model. To avoid being noticed by unauthorized users, the method ensures that the correct class probability remains dominant, preventing anomaly classifications.

\section{Related Work}
\subsection{Threat Model}
We assume the following threat model, the rightful owner is the creator of the original model, while the unauthorized user is an entity that has illicitly obtained and deployed a copied model in a cloud service which will be available as an online API. The rightful owner seeks to verify the identity of the model in two cases: (i) when attempting to confirm whether the copied model in the cloud is identical to the original model, and (ii) when proving to a third party that the copied model is identical to the original model. Since the unauthorized user denies the rightful owner's claim and is unlikely to disclose model parameters, the owner must conduct verification without direct access to the copied model's internal structure and weight parameters. Moreover, given that the unauthorized user can observe all queries and outputs, it may attempt to manipulate or block specific queries to hinder verification.
\subsection{DNN Watermarking}
DNN watermarking embeds information into models to prevent unauthorized use and theft \cite{watermark-survey}. Methods include black-box watermarking, which utilizes input-output relationships, white-box watermarking, which embeds information in model parameters, and gray-box watermarking, which embeds watermarks into probability outputs \cite{CustomizedWatermarking2022}. Backdoor-based watermarking, such as that proposed by Adi et al. \cite{adi}, modifies models to return specific outputs for trigger inputs. However, these methods face several challenges:(i) If a model is already published without embedded watermarking information, verification of ownership becomes challenging. (ii) Since watermarking techniques require models to learn information unrelated to their primary task, their outputs may exhibit recognizable patterns. These patterns can be statistically analyzed by malicious users, leading to anomaly detection. Retraining the model may further reduce the effectiveness of the watermark, diminishing its verification capabilities. (iii) The presence of hidden watermarks may be noticed and it will be removed/modified through retraining or fine-tuning \cite{chen2018, yu2023}.
Furthermore, the requirement for DNN watermarking techniques to encode task-unrelated information may lead to a decline in model accuracy.

\subsection{DNN Fingerprinting}
DNN fingerprinting extracts distinctive model characteristics to verify identity. This approach utilizes adversarial samples designed near decision boundaries and verifies whether a target model exhibits specific behavior \cite{IPGuard}. Unlike watermarking, it does not require embedding information directly into the model. However, if the fingerprinting mechanism is statistically analyzed, it may be detected as an anomaly. In particular, the model's output tendencies and decision boundary patterns can be analyzed to identify the presence of fingerprinting, leading to potential retraining or fine-tuning to circumvent verification.
\section{Proposed Method}
In this study, we propose an ownership verification method that applies adversarial attacks to enable both the owner and a third party to verify the identity of an original model and a copied model deployed in a cloud environment.

\subsection{Assumed Environment}
\begin{figure}[t]
\vspace{20pt}
\centering
\includegraphics[bb=-150 0 750 350, scale=0.3]{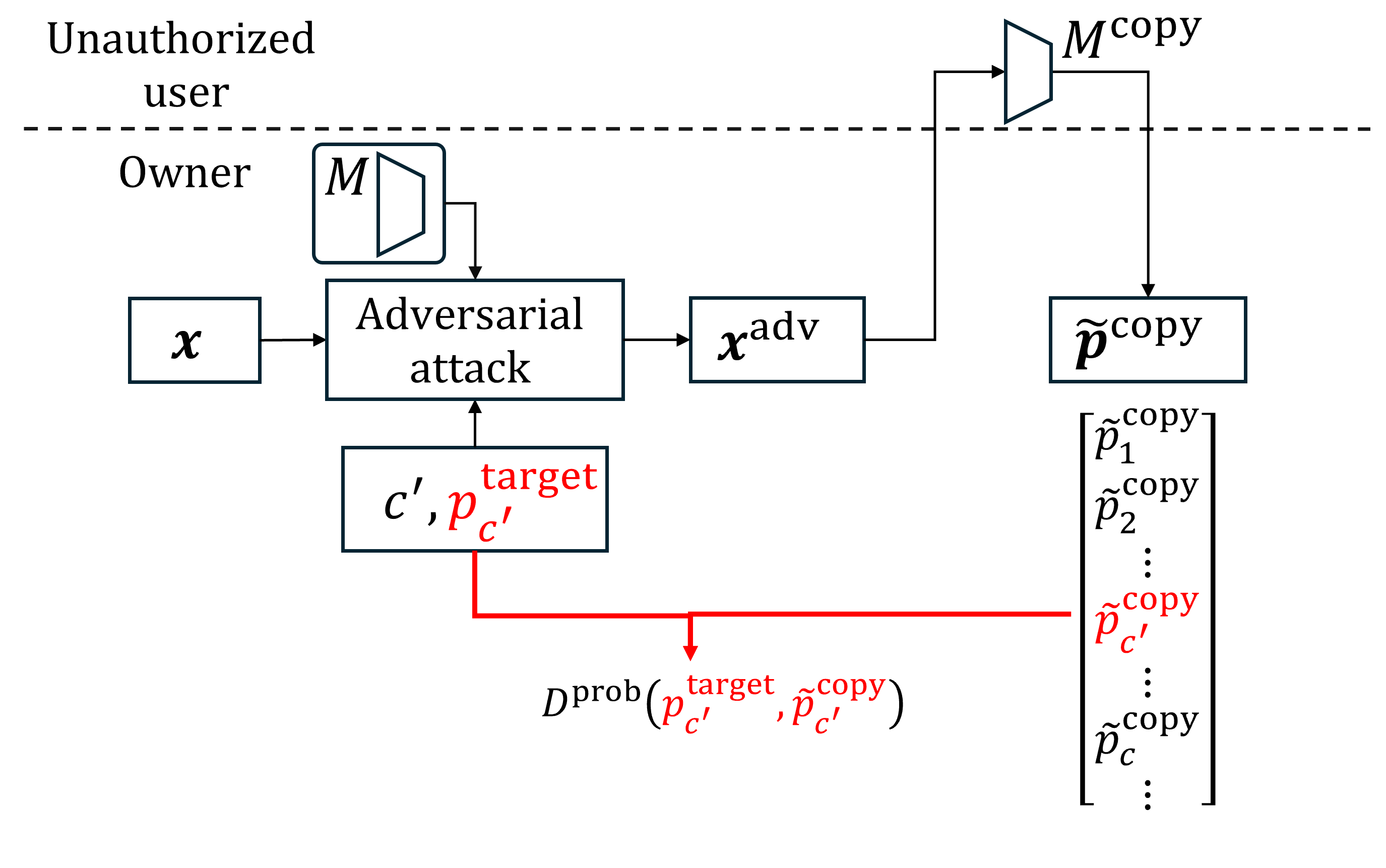}
\caption{The owner verifies the identity of the original model and the copied model.}
\label{fig:proposed_case}
\end{figure}

\begin{figure}[t]
\centering
\includegraphics[bb=-150 0 750 350, scale=0.3]{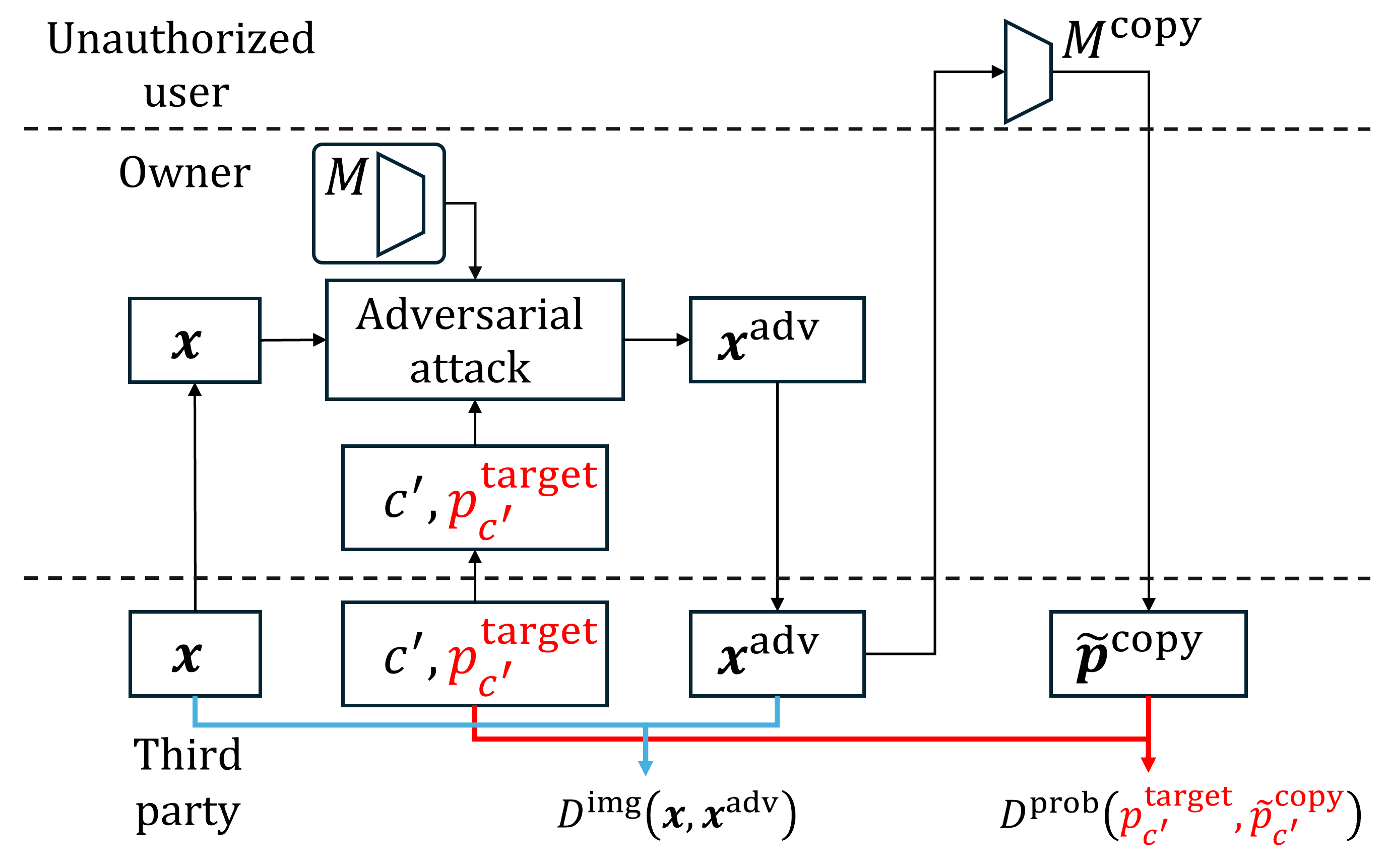}
\caption{A third party verifies the identity of the original model and the copied model by requesting the owner to create the adversarial sample for a given image with specified conditions.}
\label{fig:proposed_case2}
\end{figure}

We consider two scenarios for verifying the identity of an original model and a copied model:
(i) The owner verifies whether the copied model deployed in the cloud is identical to the original model, as illustrated in Fig.~\ref{fig:proposed_case}. (ii) The owner proves the identity of the copied model to a third party and claims ownership, as illustrated in Fig.~\ref{fig:proposed_case2}.


In Fig.~\ref{fig:proposed_case} and Fig.~\ref{fig:proposed_case2}, $M$ and $M^{\text{copy}}$ represent a classification DNN model for $k$-class image classification designed and trained by a rightful owner and its unauthorized copied model, respectively. These models do not contain special embedding structures. Model $M$ is preserved locally only at the owner, whereas $M^{\text{copy}}$ is deployed in a cloud environment and operates as a gray-box scenario such that it outputs class probability distributions.

Let $\bvec{x}$ denote an input image. The output probability vectors of $M$ and $M^{\text{copy}}$ for $\bvec{x}$ are represented as $\bvec{p}$ and $\bvec{p}^{copy}$, respectively, where each element corresponds to the probability of a class in a $k$-dimensional vector. Similarly, the output probability vectors for an adversarial sample $\bvec{x}^{\text{adv}}$ are denoted as $\bvec{\tilde{p}}$ and $\bvec{\tilde{p}}^{copy}$ for $M$ and $M^{\text{copy}}$, respectively. The correct class for $\bvec{x}$ under $M$ is $c=\underset{i}{\operatorname{argmax}}~p_i$. Additionally, let $D^{\text{prob}}(\cdot)$ and $D^{\text{img}}(\cdot)$ be functions that calculate probability distance and perceptual image distance, respectively.

\subsection{Adversarial Attack}
An adversarial attack is a technique that generates adversarial samples by intentionally perturbing input data to induce misclassification in a machine learning model. These perturbations are small enough to be imperceptible to the human eye, making it difficult to be recognized.

The computation of perturbations is categorized into two types based on access privileges:
(i) White-box attacks: Executed when the internal structure and parameters of the model are accessible.
(ii) Black-box attacks: Executed when only input-output observations are available.

Additionally, adversarial attacks are classified based on their objectives:
(i) Targeted attack: Aims to misclassify an adversarial sample into a specific class.
(ii) Untargeted attack: Aims to misclassify an adversarial sample into any incorrect class.

A targeted attack is more accurate in a white-box setting with internal model access, whereas its success rate drops in a black-box setting due to limited information.

\subsection{Overview}
The objective of the owner, as illustrated in Fig.~\ref{fig:proposed_case}, is to verify whether $M = M^{\text{copy}}$. To do this, the owner specifies a target class ${c}^\prime$ and its target probability ${p}^{\text{target}}_{c^\prime}$. Using $M$, the owner generates an adversarial sample $\bvec{x}^{\text{adv}}$ via a white-box targeted adversarial attack under the restriction of the perceptual distance between $\bvec{x}$ and $\bvec{x}^{\text{adv}}$, $D^{\text{img}}(\bvec{x}, \bvec{x}^{\text{adv}})$. Remember that $\bvec{x}^{\text{adv}}$ must not be excessively altered from the original image $\bvec{x}$. The output probability vector $\bvec{\tilde{p}} = M(\bvec{x}^{\text{adv}})$ is expected to satisfy $\tilde{p}_{c^\prime} \approx p^{\text{target}}_{c^\prime}$, while $\underset{i}{\operatorname{argmax}}~\tilde{p}_i$ remains $c$ to prevent the unauthorized user from detecting the verification approach.

Next, the adversarial sample $\bvec{x}^{\text{adv}}$ is fed into $M^{\text{copy}}$, and the output probability vector $\bvec{\tilde{p}}^{\text{copy}} = M^{\text{copy}}(\bvec{x}^{\text{adv}})$ is obtained. If the probability distance between the specified probability and the observed probability, $D^{\text{prob}}({p}^{\text{target}}_{c^\prime}, \tilde{p}^{\text{copy}}_{c^\prime})$, is sufficiently small, it determines that $M$ and $M^{\text{copy}}$ are identical.

In the scenario depicted in Fig.~\ref{fig:proposed_case2}, the owner provides proof of model identity to a third party. The third party selects an arbitrary image $\bvec{x}$, a target class $c^\prime$, and a probability $p^{\text{target}}_{c^\prime}$, and provides them to the owner as a request. The owner, possessing complete knowledge of $M$, generates an adversarial sample $\bvec{x}^{\text{adv}}$ according to the request and returns it to the third party. The third party then queries $M^{\text{copy}}$ with $\bvec{x}^{\text{adv}}$ and evaluates the response whether the observed probability satisfies $\tilde{p}^{\text{copy}}_{c^\prime} \approx p^{\text{target}}_{c^\prime}$, namely $D^{\text{prob}}(p^{\text{target}}_{c^\prime}, \tilde{p}^{\text{copy}}_{c^\prime})\approx 0$. Since accurate adversarial attacks are only possible with full model access, successful verification confirms the owner's possession of $M^{\text{copy}}$.


\subsection{Generation of Adversarial Samples}

In this study, we propose a new approach based on the targeted I-FGSM (Iterative-Fast Gradient Sign Method) \cite{I-FGSM} as a white-box adversarial attack that controls the probability of a targeted class while satisfying $c= \underset{i} {\operatorname{argmax}} \ \tilde{p}_i$. We propose targeted I-FDGSM (Iterative-Fast Dual Gradient Sign Method), a novel method that allows simultaneous control of two class probabilities.

\subsubsection{I-FGSM}
I-FGSM is a white-box adversarial attack proposed by Kurakin et al., which is an extension of FGSM \cite{FGSM}. It utilizes the gradient of the loss function obtained through backpropagation to compute perturbations iteratively. Let $\bvec{x}^{\text{adv}}_i$ represent the adversarial sample images of $\bvec{x}$ at $i$-th iteration. The update rule for targeted I-FGSM is expressed as follows:
\begin{gather}
    \bvec{x}^{\text{adv}}_0 = \bvec{x},  \notag \\ 
    \bvec{x}^{\text{adv}}_{N+1} =  \text{Clip}_{\bvec{x},\varepsilon}\left\{\bvec{x}^{\text{adv}}_{N} - \alpha^{c^\prime}\ \text{sign}\left(\nabla_x C(\bvec{x}^{\text{adv}}_{N},c^\prime\right) \right\}.\label{eq:I-FGSM}
\end{gather}

Here, $\varepsilon, \alpha^{c^\prime}, N, C$ represent the maximum perturbation range, the magnitude of perturbation at each step, the number of iterations, and the loss function, respectively. The term $\nabla_x C(\bvec{x}^{\text{adv}}{N}, c^\prime)$ denotes the gradient of the loss function $C$ concerning the target class $c^\prime$. By applying perturbations in the inverse direction of this gradient, the loss $C$ for the target class $c^\prime$ is reduced, thereby enabling a targeted attack. However, this method only considers the influence on $c^{\prime}$, leading to a situation where the probability of $c^{\prime}$ becomes excessively high while the probability of $c$ decreases significantly.

\subsubsection{I-FDGSM}
To maintain the highest probability of $c$ while adjusting the probability value of $c^{\prime}$, we propose I-FDGSM, formulated as follows:
\begin{gather}
    \bvec{x}^{\text{adv}}_0 = \bvec{x}, \notag\\
    \bvec{x}^{\text{adv}}_{N+1} = \text{Clip}_{\bvec{x},\varepsilon}\Big\{\bvec{x}^{\text{adv}}_{N} - \notag\\ \alpha^{\text{com}}\ \text{sign}\left(\beta^{c}\nabla_x C(\bvec{x}^{\text{adv}}_{N},c) + \beta^{c^{\prime}}\nabla_x C(\bvec{x}^{\text{adv}}_{N},c^{\prime})\right)\Big\}\label{eq:I-FDGSM}
\end{gather}

Here, $\alpha^{\text{com}}, \beta^c, \beta^{c^{\prime}}$ represent the magnitude of perturbation at each step and the coefficients applied to the gradients of $c$ and $c^{\prime}$, respectively. By appropriately tuning these parameters, it becomes possible to control the probabilities of both classes. It leverages the interaction between the gradients of the two classes to reduce perturbations and improve effectiveness.

Algorithm \ref{alg1} illustrates the procedure to create adversarial samples $\bvec{x}^{\text{adv}}$. 

\begin{algorithm}[t]
    \caption{I-FDGSM}
    \label{alg1}
    \begin{algorithmic}[1]
    \Require \parbox[t]{\textwidth}{Original Image: $\bvec{x}$, 
    Model: $M$, \\Target Classes: $c, c^{\prime}$, Target Probability Value: $p^{\text{target}}_{c^{\prime}}$, \\
    Factors of I-FDGSM: $\alpha^{\text{com}}, \beta^{c} = 1, \beta^{c^{\prime}} = 1$, \\Averaging Interval: $l$, Tolerance for Error: $T^{\text{diff}}$}
    \Ensure Adversarial Image: $\bvec{x}^{\text{adv}}$
    
    \State $\bvec{x}^{\text{{adv}}}_0 = \bvec{x}$
    \For{$N \in \{ 1,2,\cdots,N^{\text{max}}\}$}
        \State $\bvec{x}^{\text{{adv}}}_{N} = Adv(\bvec{x}^{\text{{adv}}}_{N-1}, M, \alpha^{\text{com}}, \beta^{c}, \beta^{c^{\prime}})$
        \State $\bvec{\tilde{p}_N} = M(\bvec{x}^{\text{{adv}}}_{N})$
        \If{$N\mod{l} = 0$}
            \State $\bvec{\tilde{p}}^{\text{mean}} = \frac{1}{l} \sum_{i=N-l+1}^{N} \bvec{\tilde{p}_i}$
            \If{$\tilde{p}^{\text{mean}}_{c^{\prime}} < (1-T^{\text{diff}})p^{\text{target}}_{c^{\prime}}$}  
                \State $\beta^{c^{\prime}} \gets \beta^{c^{\prime}}+1$
            \Else
                \State $\beta^{c} \gets \beta^{c}+1$
            \EndIf
            \If{$
                    (1-T^{\text{diff}})p^{\text{target}}_{c^{\prime}}<\tilde{p}^{\text{mean}}_{c^{\prime}} <(1+T^{\text{diff}})p^{\text{target}}_{c^{\prime}}
                $}
                \State $\alpha^{\text{com}}\gets 0.5\ \alpha^{\text{com}}$
            \EndIf
            \If{$\alpha^{\text{com}}<10^{-10}$}
                \State $\bvec{x}^{\text{adv}}=\bvec{x}^{\text{adv}}_N$
                \State \textbf{break}
            \EndIf
        \EndIf
    \EndFor
\end{algorithmic}
\end{algorithm}

\section{Computer Simulation}
\label{sec:}

In this study, both $M$ and $M^{\text{copy}}$ are set as CNN models performing an image classification task on the ImageNet dataset \cite{ImageNet}, consisting of $k=1000$ classes. In the following sections, we describe experiments evaluating the adversarial sample generation capability of I-FDGSM and the verification of model ownership.

\subsection{Adversarial Sample Generation}

We conducted a comparative experiment to evaluate the capability of adversarial sample generation using I-FGSM and I-FDGSM.

For validation, perturbations are applied to an image $\bvec{x}$ randomly selected from ImageNet based on the information of $M$. The model $M$ is set as a pre-trained ResNet50-v1 \cite{ResNet50}, provided by PyTorch \cite{pytorch}. Using Eq.\eqref{eq:I-FGSM} and Eq.\eqref{eq:I-FDGSM}, adversarial samples $\bvec{x}^{\text{adv}}$ are generated over a maximum of $N^{\text{max}}$ iterations, and the output $\bvec{\tilde{p}}=M(\bvec{x}^{\text{adv}})$ is obtained for each iteration $N$. The variations of $\tilde{p}_c$ and $\tilde{p}_{c^\prime}$ with respect to $N$ are analyzed.

The parameters for each method are set as follows for validation: $N^{\text{max}} = 1000$, $\alpha^{c^\prime}=5\times10^{-4}$, $\alpha^{\text{com}}=1\times10^{-3}$, $\varepsilon = 0.05$, $l = 5$, and $T^{\text{diff}}=5\times10^{-3}$. These parameters were chosen empirically to ensure stable convergence and effective probability control in preliminary tests.

\subsubsection{Results and Discussion}

\begin{figure}[b]
\vspace{55pt}
\centering
\begin{minipage}[b]{0.49\columnwidth}
    \centering
    \includegraphics[bb=15 0 700 100, scale=0.23]{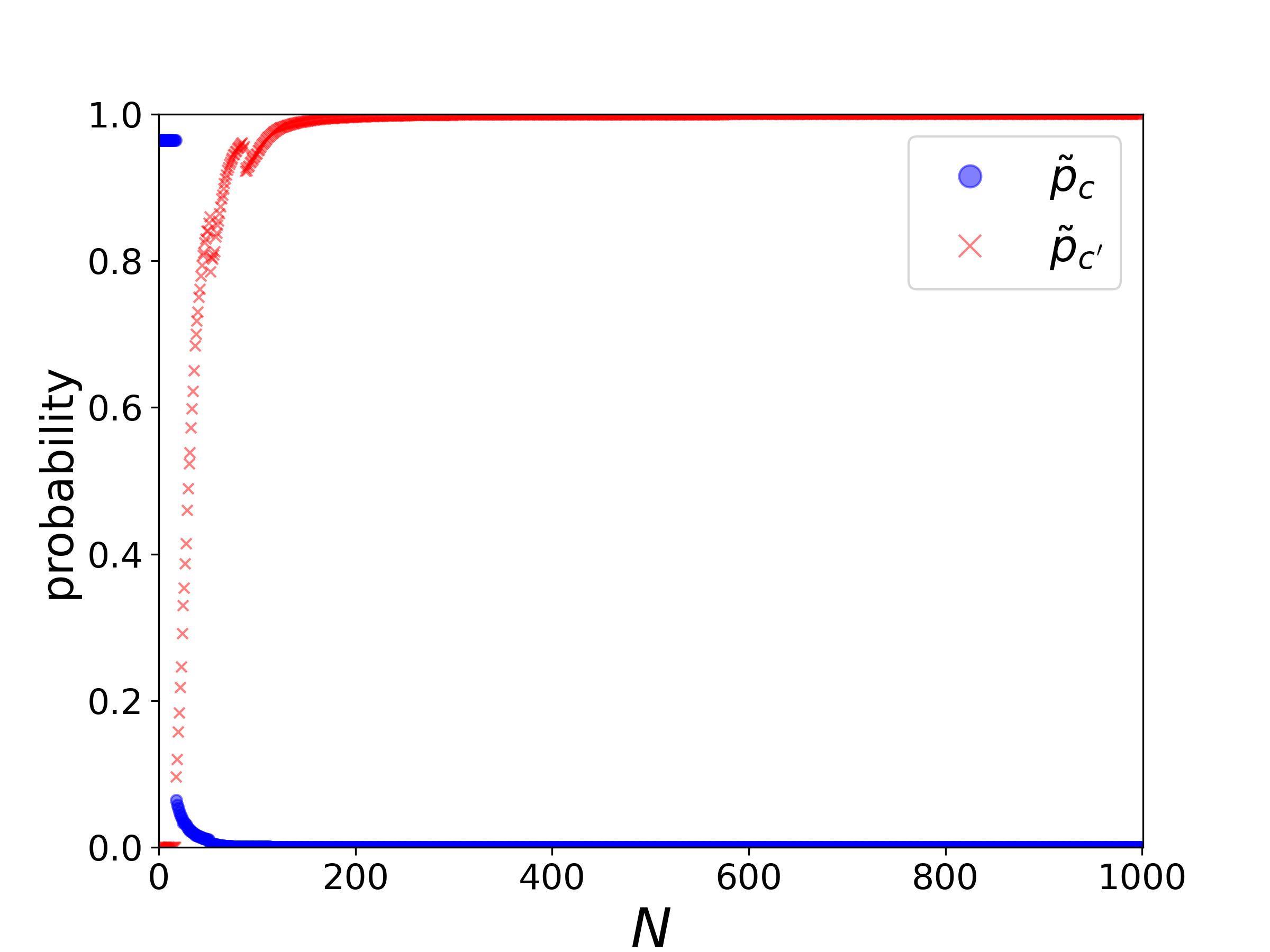}
    \smallskip
    \text{(a) I-FGSM}
\end{minipage}
\begin{minipage}[b]{0.49\columnwidth}
    \centering
    \includegraphics[bb=0 0 700 100, scale=0.23]{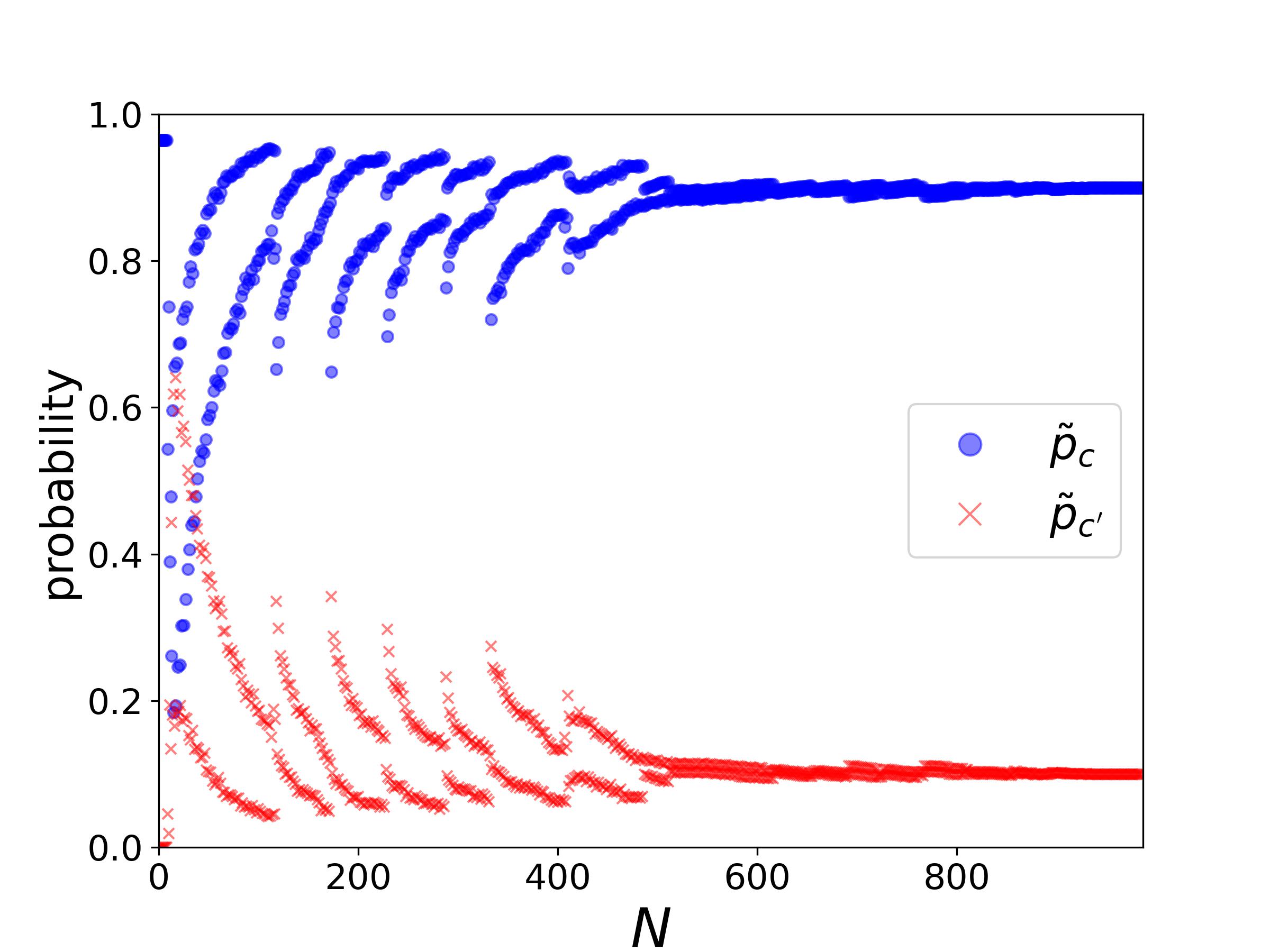}
    \smallskip
    \text{(b) I-FDGSM, $p^{\text{target}}_{c^{\prime}}=0.1$}
\end{minipage}

\vspace{75pt} 

\begin{minipage}[b]{0.49\columnwidth}
    \centering
    \includegraphics[bb=0 0 700 100, scale=0.23]{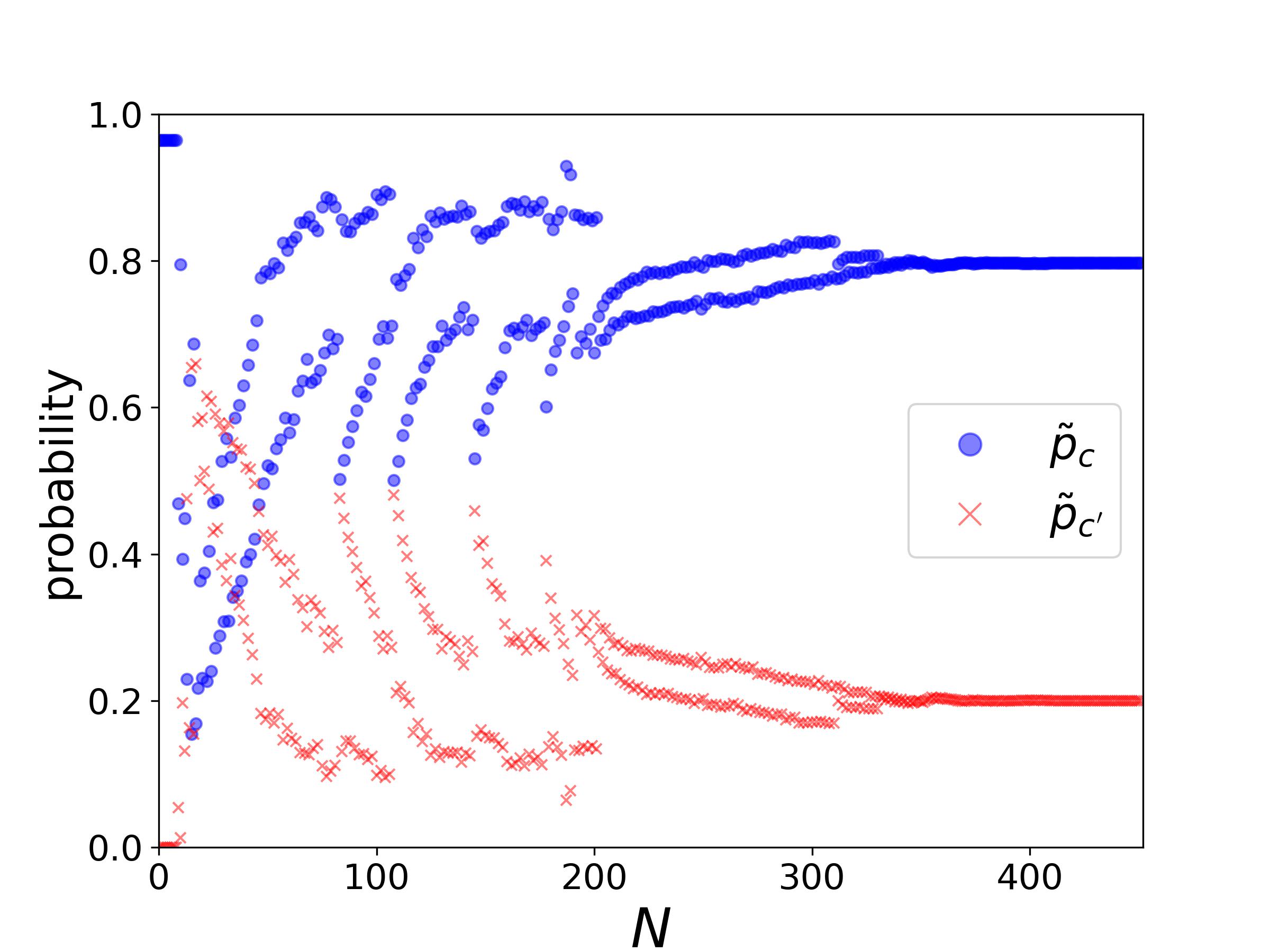}
    \smallskip
    \text{(c) I-FDGSM, $p^{\text{target}}_{c^{\prime}}=0.2$}
\end{minipage}
\begin{minipage}[b]{0.49\columnwidth}
    \centering
    \includegraphics[bb=0 0 700 100, scale=0.23]{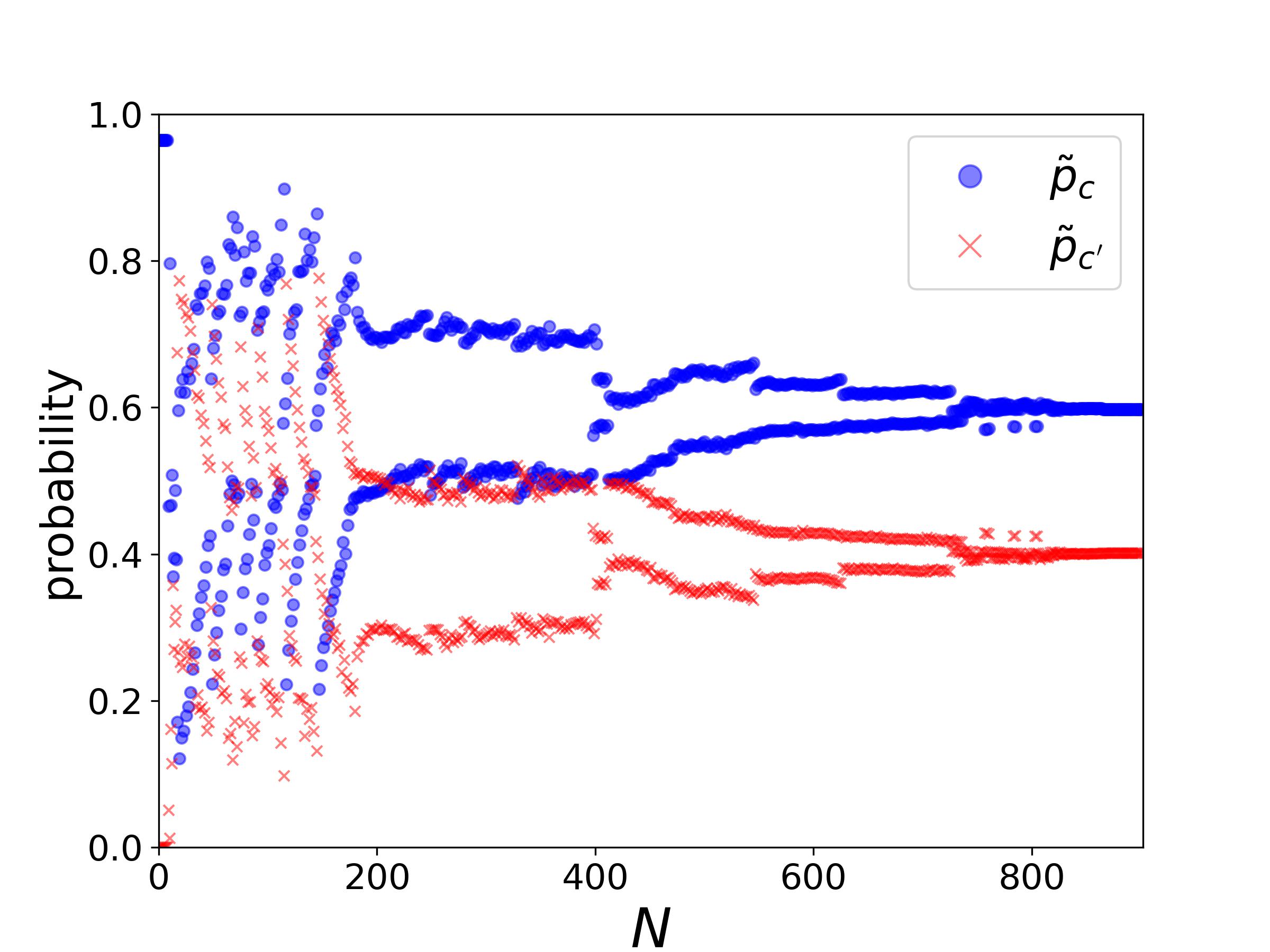}
    \smallskip
    \text{(d) I-FDGSM, $p^{\text{target}}_{c^{\prime}}=0.4$}
\end{minipage}

\caption{Variations of $\tilde{p}_{c}$ and $\tilde{p}_{c^\prime}$ with respect to $N$. (a): I-FGSM, (b)(c)(d): I-FDGSM, with $p^{\text{target}}_{c^{\prime}}=0.1, 0.2, 0.4$.}
\label{fig:ex1}
\end{figure}

The results of I-FGSM and I-FDGSM are shown in Fig.~\ref{fig:ex1}. The blue circles and red crosses in the figure represent the variations of $\tilde{p}_{c}$ and $\tilde{p}_{c^\prime}$ with respect to the number of iterations $N$.

From Fig.~\ref{fig:ex1} (a), in I-FGSM, $\tilde{p}_c$ and $\tilde{p}_{c^{\prime}}$ rapidly decrease and increase, respectively, as the number of iterations $N$ increases, with $\tilde{p}_{c^{\prime}}$ converging to a value close to 1. On the other hand, from Fig.~\ref{fig:ex1} (b)(c)(d), in I-FDGSM, $\tilde{p}_{c^{\prime}}$ converges to a value sufficiently close to the specified $p^{\text{target}}_{c^{\prime}}$. Furthermore, $\tilde{p}_c$ remains at the highest probability value and converges with the increase of $N$. It is interesting to note that the sum of $\tilde{p}_c$ and $\tilde{p}_{c^{\prime}}$ approaches 1. This result suggests that, in Eq.~\eqref{eq:I-FDGSM}, considering the gradient influences of both $c$ and $c^{\prime}$ enables the adjustment of probabilities to the local optimal values for the two classes.

\subsection{Ownership Verification of Models}
To verify the identity of copied models using I-FDGSM, we conducted ownership verification experiments.

\subsubsection{Experimental Conditions}

The input image $\bvec{x}$ is randomly selected from the ImageNet dataset, and $\bvec{x}^{\text{adv}}$ is generated using I-FDGSM to satisfy $\tilde{p}_{c^{\prime}} \approx p^{\text{target}}_{c^{\prime}}$ based on Eq \eqref{eq:I-FDGSM} and Algorithm \ref{alg1}. In practice, if the owner performs the verification, $c^\prime$ and $p_{c^\prime}$ are specified by the owner. Meanwhile, when a third party conducts the verification, $\bvec{x}$ and $c^\prime$ and $p_{c^\prime}$ are specified by the third party.

The probability distance function $D^{\text{prob}}(\cdot)$ is used to calculate the relative error between the obtained output probability $\tilde{p}^{\text{copy}}_{c^\prime}$ and the target probability $p^{\text{target}}_{c^\prime}$, as given by Eq.~\eqref{eq:RE}:
\begin{equation}
D^{\text{prob}}(p^{\text{target}}_{c^\prime},\tilde{p}^{\text{copy}}_{c^\prime})=\frac{|p^{\text{target}}_{c^\prime}-\tilde{p}^{\text{copy}}_{c^\prime}|}{p^{\text{target}}_{c^\prime}}.
\label{eq:RE}
\end{equation}

Furthermore, the perceptual distance function $D^{\text{img}}(\cdot)$ is used to measure the similarity between $\bvec{x}$ and $\bvec{x}^{\text{adv}}$ using $\operatorname{SSIM}(\cdot)$ (Structural Similarity Index Measure) \cite{SSIM}, which considers human visual perception.

The simulation conditions are summarized in Table \ref{tab:ex2_condition}.

\begin{table}[tbp]
    \caption{Simulation Conditions}
    \centering
    \begin{tabular}{cc}
        \hline
        $c^\prime$ & Randomly selected from 1000 classes \\ 
        $p^{\text{target}}_{c^\prime}$ & Uniform\{0.1, 0.15, 0.2, 0.25, 0.3, 0.35, 0.4\} \\ 
        $ \alpha^{\text{com}}, \varepsilon, l, T^{\text{diff}}$ & $1\times10^{-3}, 5\times10^{-2}, 5, 5\times10^{-3}$ \\
        \hline
    \end{tabular}
    \label{tab:ex2_condition}
\end{table}

\subsubsection{Results and Discussion}

\begin{figure}[tbh]
\vspace{10pt}
    \centering
    \includegraphics[bb=125 0 500 350, scale=0.3]{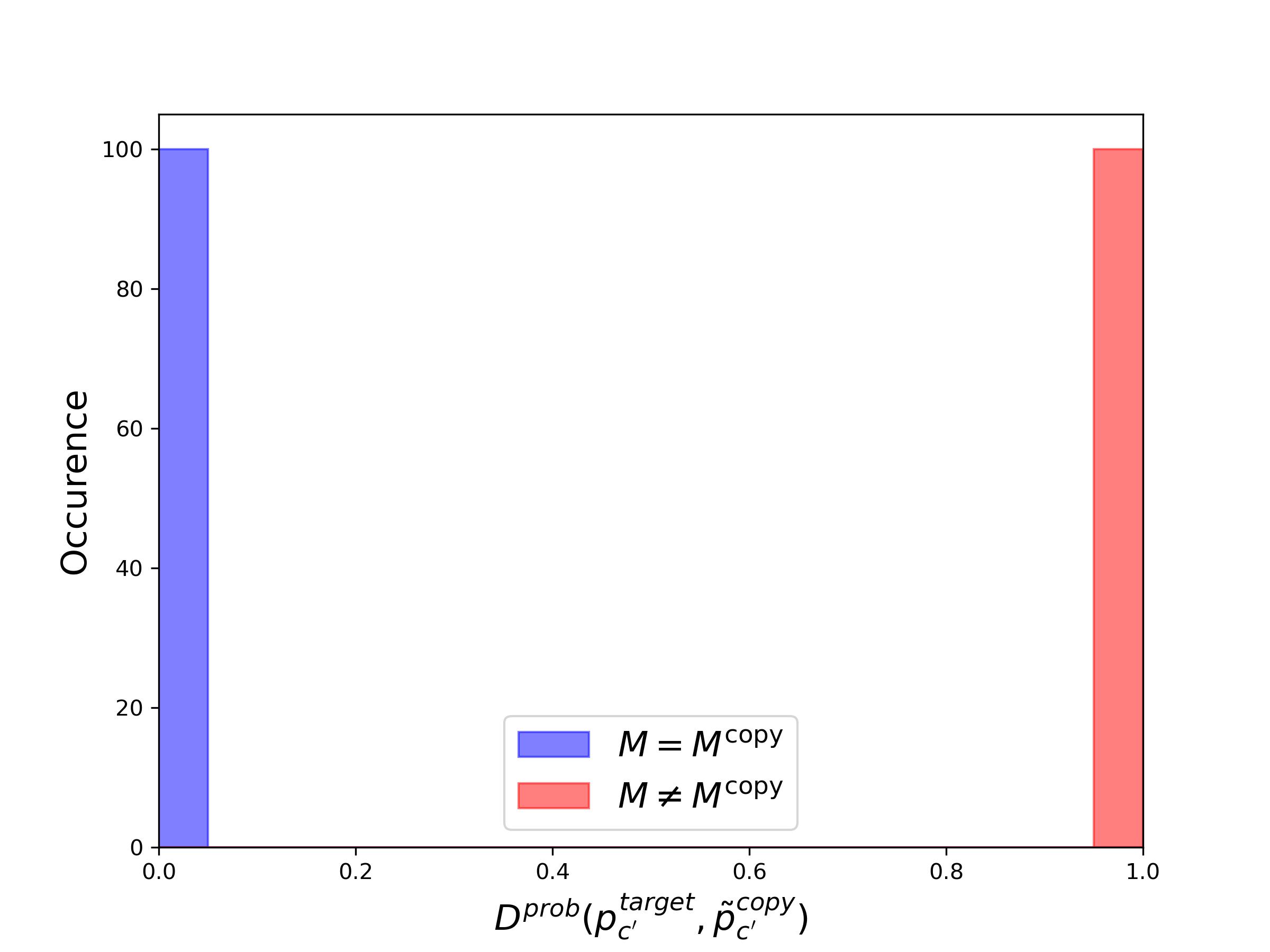}
    \caption{Distribution of $D^{\text{prob}}(p^{\text{target}}_{c^\prime},\tilde{p}^{\text{copy}}_{c^\prime})$. Blue bars: $M = M^{\text{copy}} =$ ResNet50-v1. \ Red bars: $M =$ ResNet50-v1, $M^{\text{copy}} =$ ResNet50-v2.}
\label{fig:ex2_RE_hist_Res50v1_Res50v2}
\end{figure}

Using PyTorch's ResNet50-v1 as $M$ and ResNet50-v1 and ResNet50-v2 as $M^{\text{copy}}$, we evaluated 100 randomly selected images from ImageNet. The results are shown in Fig.~\ref{fig:ex2_RE_hist_Res50v1_Res50v2}, where blue bars represent the cases where $M = M^{\text{copy}}$, and red bars represent the cases where $M \neq M^{\text{copy}}$.

As shown in Fig.~\ref{fig:ex2_RE_hist_Res50v1_Res50v2}, $D^{\text{prob}} \approx 0$ for $M = M^{\text{copy}}$ and $D^{\text{prob}}\approx 1$ otherwise. This result indicates that adversarial samples generated using I-FDGSM are well adjusted for $M$, allowing more accurate probability control only for the model. Consequently, if $\tilde{p}_{c^{\prime}}$ in $M^{\text{copy}}$ can be controlled, it indicates that the owner has sufficient knowledge of $M^{\text{copy}}$, proving the ownership of $M^{\text{copy}}$ to a third party without presenting it.

\begin{figure}[b]
\vspace{15pt}
    \centering
    \includegraphics[bb=0 0 500 350, scale=0.4]{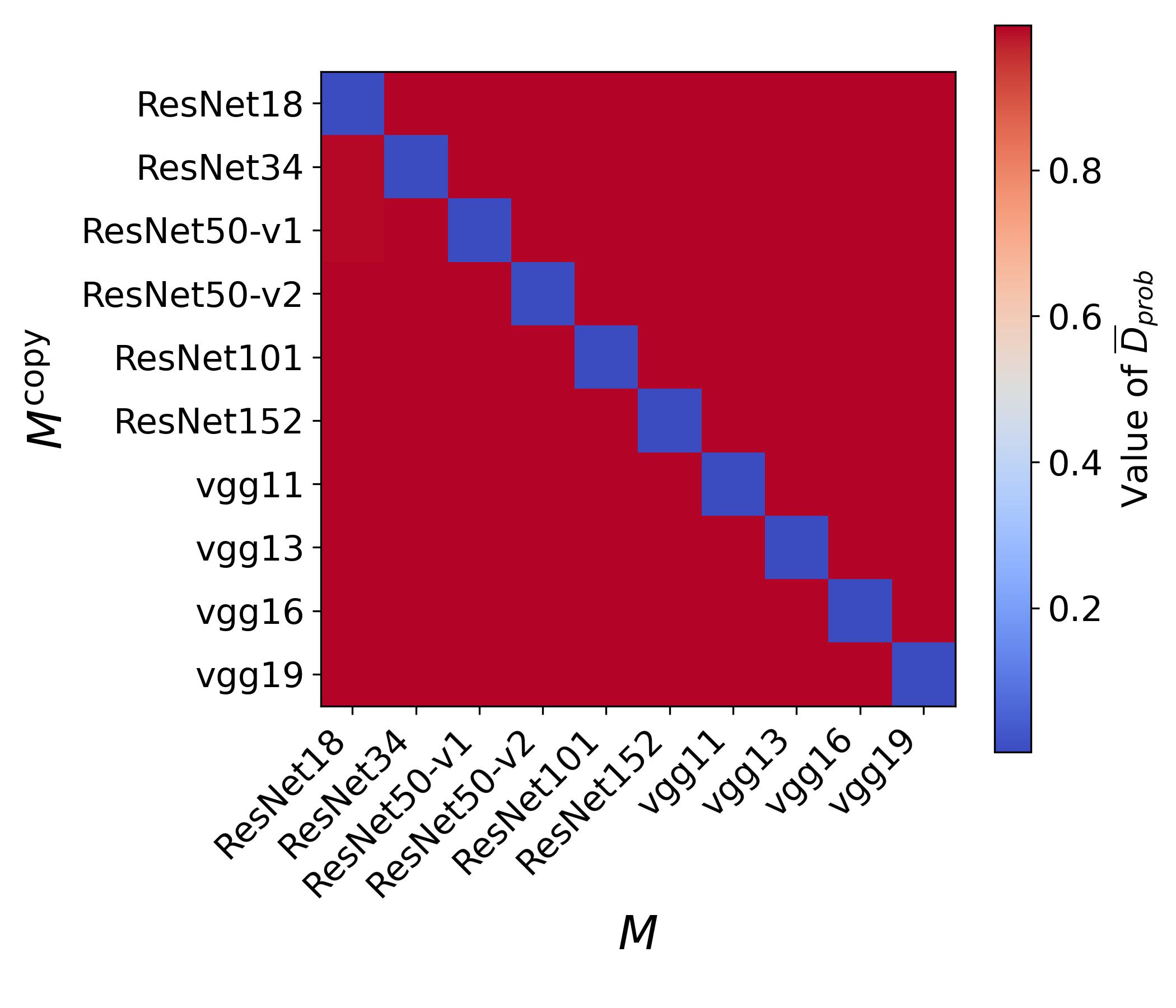}
    \caption{Heatmap of the average $D^{\text{prob}}(p^{\text{target}}_{c^\prime},\tilde{p}^{\text{copy}}_{c^\prime})$ values for multiple $M$ and $M^{\text{copy}}$.}
\label{fig:ex2_ManyModels_RE_ave}
\end{figure}

Next, model ownership verification was conducted using various ResNet and VGG models\cite{VGG} from PyTorch. For each model, 100 randomly selected images from ImageNet were used as $\bvec{x}$. 
Fig.~\ref{fig:ex2_ManyModels_RE_ave} represents the average of $D^{\text{prob}}(p^{\text{target}}_{c^\prime},\tilde{p}^{\text{copy}}_{c^\prime})$ value, denoted by $\overline{D}^{\text{prob}}$, for each pair. From the figure, when $M = M^{\text{copy}}$, $\overline{D}^{\text{prob}} \approx 0$, and when $M \neq M^{\text{copy}}$, $\overline{D}^{\text{prob}} \approx 1$. This indicates that probability value adjustments are possible only for specific models, thereby proving the ownership of $M^{\text{copy}}$ across all tested model pairs. 

It is worth noting that the average of $\operatorname{SSIM}(\bvec{x},\bvec{x}^{\text{adv}})$ calculated for each of the 100 generated images per $M$ was at least 0.9875. This confirms that perturbations introduced by I-FDGSM are visually imperceptible, making detection by unauthorized users difficult.

\section{Conclusion and Future Work}

In this study, we proposed a novel framework for verifying the identity of trained DNN models without presenting the original model. This framework enables the rightful owner to prove the model's identity to a third party while preserving confidentiality. To avoid detection by unauthorized users, we introduced I-FDGSM, an adversarial attack method that precisely controls the probability values between the original and target classes. Experiments confirmed its high-accuracy verification capability.

Due to the transferability of adversarial perturbations, the proposed method is expected to remain effective against slightly modified models such as those retrained or pruned. Evaluating this robustness is left for our future work. Furthermore, one of the threats in the proposed method is the potential detectability by anomaly detection mechanisms. A prior work has shown that such patterns can be flagged even in encrypted domains like VoIP traffic \cite{ADVoIP}. Improving stealth and robustness against such detection is another direction for future work.



\bibliographystyle{IEEEtran}
\bibliography{refs}

\end{document}